%% file: SensorBasedLTL_SemanticMaps_Apr10.tex
\documentclass[letterpaper, 10 pt, conference]{ieeeconf}

\IEEEoverridecommandlockouts                              
\newtheorem{theorem}{Theorem}[section]

\newtheorem{problem}{Problem}
\newtheorem{definition}[theorem]{Definition}
\newtheorem{rem}[theorem]{Remark}

\usepackage{graphicx} 
\usepackage{amsmath}
\usepackage{amssymb}
\usepackage{epstopdf}
\usepackage{cite}
\usepackage[noend,ruled,linesnumbered]{algorithm2e}
\SetKwComment{Comment}{$\triangleright$\ } {}
\usepackage{multirow}
\usepackage{rotating}
\usepackage{subfigure} 
\usepackage{color} 
\usepackage{mysymbol}
\usepackage[dvipsnames]{xcolor}
\usepackage[hyphens]{url}

\usepackage[breaklinks=true, colorlinks, bookmarks=true, citecolor=Black, urlcolor=Violet,linkcolor=Black]{hyperref}

\newcommand{\set}[1]{\left\{#1\right\}}

\usepackage{comment}
\usepackage[colorinlistoftodos,prependcaption,textwidth=1.5cm,textsize=tiny]{todonotes}
\begin{document}
\title{\LARGE \bf Multi-robot Mission Planning in Dynamic Semantic Environments}
\author{Samarth Kalluraya$^1$, George J. Pappas$^2$, Yiannis Kantaros$^1$
\thanks{$^{1}$Authors are with the Department of Electrical and Systems Engineering, Washington University at St. Louis, St. Louis, MO, 63130, USA. 
        {\tt\small {k.samarth,ioannisk@wustl.edu}}}%
\thanks{$^{2}$Author is with with the GRASP Laboratory, University of Pennsylvania, Philadelphia, PA, 19104, USA. 
        {\tt\small pappasg@seas.upenn.edu}}%
}

\maketitle 
\begin{abstract}
This paper addresses a new semantic multi-robot planning problem in uncertain and dynamic environments. 
Particularly, the environment is occupied with mobile and uncertain semantic targets. 
These targets are governed by stochastic dynamics while their current and future positions as well as their semantic labels are uncertain.
Our goal is to control mobile sensing robots so that they can accomplish collaborative semantic tasks defined over the uncertain current/future positions and semantic labels of these targets. We express these tasks using Linear Temporal Logic (LTL). We propose a sampling-based approach 
that explores the robot motion space, the mission specification space, as well as the future configurations of the semantic targets to design optimal paths. These paths are revised online to adapt to uncertain perceptual feedback. To the best of our knowledge, this is the first work that addresses semantic mission planning problems in uncertain and dynamic semantic environments. We provide extensive experiments that demonstrate the efficiency of the proposed method.
\end{abstract}
\IEEEpeerreviewmaketitle
   
\section{Introduction} \label{sec:Intro}
\input{files/Intro}

\section{Problem Definition} \label{sec:PF}
\input{files/PF_ICRA}

\section{Safe Planning In Dynamic Semantic Maps}\label{sec:samplingAlg}
\input{files/alg_ICRA}
\section{Experimental Validation} \label{sec:Sim}
\input{files/sim_new}

\section{Conclusion} \label{sec:Concl}
This paper addressed a multi-robot planning problem in uncertain and dynamic semantic environments. We proposed a sampling-based algorithm to design paths that are revised online based on perceptual feedback. We validated the proposed algorithm in complex semantic navigation tasks.

\bibliographystyle{IEEEtran}
\bibliography{SK_bib.bib}

\end{document}

%% file: files/Intro.tex
Robot navigation has received considerable research attention \cite{lavalle2006planning,mohanan2018survey,elbanhawi2014sampling}. Typically, motion planning problems require generating trajectories that reach known goal regions while avoiding known/unknown, and possibly dynamic, obstacles. Recent advances in computer vision and semantic mapping offer a unique opportunity to transition from these well-studied \textit{geometric} planning approaches to \textit{semantic} mission planning problems requiring reasoning about both the geometric and semantic environmental structure \cite{rosinol2020kimera,bowman2017probabilistic,vasilopoulos2020reactive}.

In this paper we address a new semantic multi-robot planning problem in uncertain and dynamic environments. 
The environment is assumed to have known and static geometry (e.g., walls) but it is occupied with  mobile and uncertain labeled targets of interest (e.g., pedestrians, drones, etc) that do not interact with our multi-robot system. Particularly, the targets move as per known dynamics but they are subject to exogenous disturbances (e.g., wind gusts) resulting in positional uncertainty (\textit{metric} uncertainty).
The labels of these targets are also initially unknown (\textit{semantic} uncertainty). Instead, the robots have access to a probabilistic prior belief about the initial positions and the semantic labels of the targets. This prior belief may be user-specified or computed by existing semantic mapping methods \cite{rosinol2020kimera,bowman2017probabilistic}. The goal of the robots is to accomplish collaborative \textit{semantic tasks} defined over the uncertain positions and/or the semantic labels of the targets. These tasks are expressed using Linear Temporal Logic (LTL) \cite{baier2008principles}.
%
To accomplish them, the robots are equipped with imperfect perception systems (e.g., cameras and learning-based object detectors) that allow them to reason about the semantic environmental structure by detecting, classifying, and localizing objects. The considered planning problem gives rise to an optimal control problem that generates open-loop control policies. 
To solve this new problem, building upon our previous work \cite{Kantaros2020perception}, we propose a sampling-based approach that explores the robot motion space, the mission specification space, as well as the uncertain future states of the mobile semantic targets. The offline designed control policies are updated online to adapt to uncertain perceptual feedback. 
Extensions to targets with fully unknown dynamics are also discussed.  


\begin{figure}[t]
  \centering
\includegraphics[width=1\linewidth]{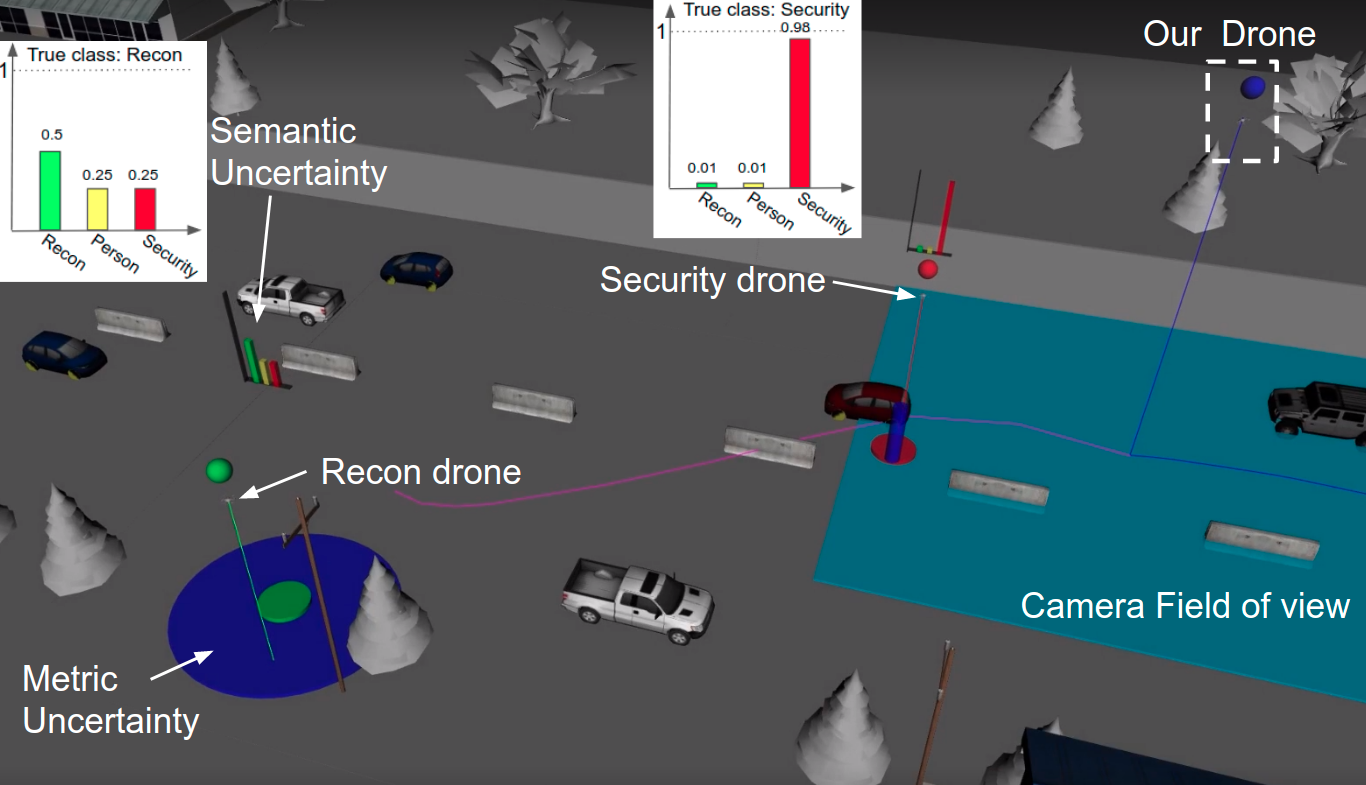}
\caption{\textcolor{black}{A drone equipped with a noisy downward facing camera (blue square), and an object detector is responsible for taking photos of moving targets of interest  with semantic labels `recon drone' and `person', while always avoiding flying close to targets with semantic labels `enemy security drone'. The actual semantic labels and the locations of these targets as well as their dynamics are uncertain. The blue ellipsoids and histograms capture positional/metric and semantic uncertainty of the targets, respectively. Recon and security drones are depicted with green and red spheres, respectively. 
}} 
  \label{fig:intro_fig}
\end{figure}

\textbf{Related works:} Several motion planning algorithms have been proposed that assume known  \cite{karaman2011sampling,kavraki1996probabilistic,kantaros2018text,leahy2016persistent,fainekos2005temporal,kloetzer2010automatic} or unknown \cite{vasilopoulos2018sensor,bansal2019combining,ryll2019efficient,zhai2019path,elhafsi2019map,guo2013revising,guo2015multi,maly2013iterative,lahijanian2016iterative,livingston2012backtracking,livingston2013patching,kantaros2020reactive,vasilopoulos2021reactive} but static environments. 
%
Recently, these methods have been extended to dynamic environments for both reach-avoid
\cite{guzzi2013human,otte2016rrtx,pierson2019dynamic,aoude2013probabilistically,fisac2019hierarchical,wang2019game, schmerling2018multimodal, bandyopadhyay2013intention, fridovich2020confidence, tordesillas2021panther,cleaveland2022learning}
and temporal logic tasks \cite{ulusoy2014receding, hoxha2016planning, maity2015motion, li2021safe, li2022online}. Common in these works is that they consider environments with unknown geometry but with known semantic structure. As a result, they consider tasks requiring reaching perfectly known and static regions while avoiding dynamic known/unknown obstacles. To the contrary, here, we consider semantic tasks that require reasoning about both the uncertain current/future positions and the semantic labels of mobile targets. 
For instance, consider a semantic task requiring a drone to take a picture of an `abandoned car' while avoiding `patrolling drones' where the current/future locations and the semantic labels of these targets (i.e., cars and drones) are uncertain. 
%
%
Related are also the works on active sensing for target tracking that require to actively decrease uncertainty of all mobile targets; see e.g., \cite{schlotfeldt2018anytime,atanasov2015decentralized,kantaros2021sampling,charrow2014approximate,dames2017detecting,hollinger2014sampling,bircher2017incremental,tzes2021distributed,placed2022survey}. Nevertheless, the control objective in our work is fundamentally different; our goal is to design informative paths that satisfy temporal logic missions defined over these uncertain labeled targets. To the best of our knowledge, the most relevant work to the one proposed here is the recent work by the authors \cite{Kantaros2020perception}, which, unlike this work, considers \textit{static} semantic environments. In this paper, we extend \cite{Kantaros2020perception} to dynamic environments.
\textbf{Contributions:} 
\textit{First}, we  formulate a new semantic mission planning problem in dynamic and  uncertain semantic environments.
\textit{Second}, we present a sampling-based approach to design paths that satisfy semantic missions captured by LTL specifications. 
\textit{Third}, we provide extensive experiments that demonstrate the efficiency of the proposed algorithm. 

%% file: files/PF_ICRA.tex
\subsection{Modeling Uncertain \& Dynamic Semantic Environments}\label{sec:uncertainMap}
\vspace{-0.1cm}
We consider semantic environments, denoted by $\Omega$, with known geometric structure (e.g., walls or buildings). This known obstacle-free space, denoted by $\Omega_{\text{free}}\subseteq\Omega$, is cluttered with $M>0$ mobile labeled targets $\ell_i$ giving rise to a dynamic semantic map $\ccalM(t)=\{\ell_1,\ell_2,\dots,\ell_M\}$. Each target $\ell_i=\{\bbx_i(t),c_i,\bbg_i\}\in\ccalM$ is defined by its state $\bbx_i(t)$ (e.g., position and orientation) at a time $t$, it's class $c_i\in\ccalC$, where $\ccalC$ is a finite set of classes (e.g., `car', `pedestrian', and `drone'), and dynamics $\bbg_i$. Specifically, each target is governed by the following dynamics: 
$\bbx_i(t+1)=\bbg_i(\bbx_i(t),\boldsymbol\mu_i(t),\boldsymbol\nu_i(t))$, where $\boldsymbol\mu_i(t)$ is the control input selected by target $i$ at time $t$ and $\boldsymbol\nu_i(t)$ models noise and exogenous disturbances in the target dynamics. We assume that the noise follows a Gaussian distribution, i.e., $\boldsymbol\nu_i\sim\ccalN(0,\bbR_i)$ with known covariance matrix $\bbR_i$. \textbf{Assumption (a):} Hereafter, we assume that $\bbg_i$ models linear system dynamics, i.e., $\bbx_i(t+1)=\bbA_i\bbx_i(t)+\bbB_i\boldsymbol\mu_i(t)+\boldsymbol\nu_i$. We compactly denote the dynamics of all targets by:
\begin{equation}
\bbx(t+1)=\bbA\bbx(t)+\bbB\boldsymbol\mu(t)+\boldsymbol\nu,
\end{equation}
where $\bbx$ is a vector that stacks the states of all targets. 
\textbf{Assumption (b):} We assume that the dynamics $\bbg_i$, i.e., the matrices $\bbA_i$, $\bbB_i$, $\bbR_i$ as well as $\boldsymbol\mu_i(t)$ are known for all $t$; for instance, target dynamics can be learned using existing data-driven methods. The control inputs $\boldsymbol\mu_i(t)$ are pre-determined (offline) and they are not affected by the multi-robot system introduced in Section \ref{sec:modelSensingRobot}.

The true state $\bbx_i(t)$ of target $i$ at time $t$ is uncertain. \textbf{Assumption (c):} We assume that we have access to  a prior Gaussian belief about the initial state of each target, i.e., $\bbx_i(0)\sim\ccalN(\hat{\bbx}_i(0),\Sigma_i(0))$, where  $\hat{\bbx}_i(0)$ and $\Sigma_i(0)$ denote the mean and covariance matrix for $\bbx_i(0)$.  %
We note that assumptions (a)-(c) are quite common in the related literature \cite{atanasov2015decentralized, kantaros2021sampling,placed2022survey}; in Section \ref{sec:extensions}, we discuss data-driven methods to relax these assumptions. 
As it will be discussed later, these assumptions allow us to model the uncertain future states $\bbx_i(t)$ as Gaussian distributions, using a Kalman filter (KF) approach. 
These Gaussian distributions model \textit{metric} uncertainty in the environment. Also, the true labels/classes $c_i$ of the targets $\ell_i$ are uncertain as well. \textbf{Assumption (d):}  We assume that we have access to an arbitrary discrete distribution $d_i$ for all targets $\ell_i$ that model the probability that the class for $\ell_i$ is $c_i\in\ccalC$, i.e., $c_i\sim d_i$. The discrete distributions model \textit{semantic} uncertainty in the environment. This prior  information can be provided by semantic mapping methods \cite{rosinol2020kimera,bowman2017probabilistic} or it can be user-specified.

\vspace{-0.2cm}
\subsection{Modeling Perception-based Robots}\label{sec:modelSensingRobot}
\vspace{-0.1cm}
Consider $N$ mobile robots governed by the following dynamics: $\bbp_{j}(t+1)=\bbf_j(\bbp_{j}(t),\bbu_{j}(t))$,
for all $j\in\{1,\dots,N\}$, where $\bbp_{j}(t)\in\mathbb{R}^n$ stands for the state (e.g., position and orientation) of robot $j$ in the free space $\Omega_{\text{free}}$ at discrete time $t$, and $\bbu_{j}(t) \in\mathit{U}_j$ denotes a control input selected from a \textit{finite} space of admissible controls $\mathit{U}_j$. For simplicity, we also denote $\bbp_{j}(t)\in\Omega$.
Hereafter, we compactly denote the dynamics of all robots as 
\begin{equation}\label{eq:rdynamics}
\bbp(t+1)=\bbf(\bbp(t),\bbu(t)),
\end{equation}
where $\bbp(t)\in \Omega^N$, $\forall t\geq 0$, and $\bbu(t)\in\mathit{U}:=\mathit{U}_1\times\dots\times\mathit{U}_N$.

The robots are equipped with sensors (e.g., cameras) to collect measurements associated with $\bbx(t)$. \textbf{Assumption (e):} We assume that these sensors can be modeled as per the following \textit{linear observation model}: $\bby_j(t) = \bbM_j(\bbp_j(t))\bbx(t) + \bbv_j(t)$, where $\bby_j(t)$ is the measurement signal at discrete time $t$ taken by robot $j$. Also, $\bbv_j(t) \sim \ccalN(\bb0, \bbQ_j)$ is Gaussian noise with known covariance $\bbQ_j$; similar sensor models are used e.g., in \cite{freundlich2018distributed}. We compactly denote all observation models as 
\begin{equation}\label{eq:measModelR}
\bby(t)= \bbM(\bbp(t))\bbx+\bbv(t),~\bbv(t) \sim \ccalN(\bb0, \bbQ),
\end{equation}
where $\bby(t)$ collects measurements taken at time $t$ by all robots  associated with any target. 
The robots are also equipped with object recognition systems allowing them to collect measurements associated with the target classes. These measurements typically consist of label measurements along with label probabilities \cite{redmon2016you,ren2016faster, guo2017calibration}. 
The semantic measurement is generated by the following observation model: $[\bby_{j}^c, \bbs_{j}^c]=\bbg_j(L_i)$ (see also \cite{atanasov2016localization}) where $\bby_{j}^c$ and $\bbs_{j}^c$ represent a class/label measurement and the corresponding probability scores over all available classes, respectively, and $L_i$ stands for the true class of the detected target $\ell_i$.  We compactly denote the object recognition model of all robots as  
\begin{equation}\label{eq:measModelC}
[\bby^c(t), \bbs^c(t)]=\bbg(\bbL),
\end{equation}
where $\bbL$ denote the true classes of all targets detected.
\vspace{-0.2cm}
\subsection{Kalman Filter for Offline Map Prediction}\label{sec:updateMap}
\vspace{-0.1cm}
Our belief about the metric/semantic environmental structure can be updated online by using the observations generated by \eqref{eq:measModelR}-\eqref{eq:measModelC} using existing semantic mapping methods \cite{rosinol2020kimera,bowman2017probabilistic}. Here we leverage a Kalman Filter (KF) to approximately predict offline (i.e., without observations) the future target states as well as their associated metric uncertainty. First, due to assumptions (b)-(c), the expected states of the targets at time $t+1$ can be predicted by applying recursively the KF prediction formula, i.e., $\hat{\bbx}(t+1)=\bbA\hat{\bbx}(t)+\bbB\boldsymbol\mu(t)$. Note that this is the \textit{a-priori} state estimate of targets $\ell_i$; the \textit{a-posteriori} state estimates require observations that are not available offline. Due to assumptions (a)-(c) and (e), the associated covariance matrix $\Sigma(t+1)$, capturing metric uncertainty, can be computed using the KF  Riccati equation $\Sigma(t+1)=\rho(\Sigma(t),\bbp(t+1))$. Note that this is the \textit{a-posteriori} covariance matrix which can be computed optimally  without the need for measurements \cite{atanasov2015decentralized}. Finally, the semantic uncertainty, captured by the discrete distribution $d$, cannot be updated offline as it requires field measurements (i.e., images). Hereafter, we compactly denote by $\hat{\ccalM}(t)=(\hat{\bbx}(t),\Sigma(t), d(t))$, our offline estimate for all targets at $t$. 

\vspace{-0.2cm}
\subsection{Semantic Mission \& Safety Specifications using LTL}\label{sec:task}
\vspace{-0.1cm}
The goal of the robots is to accomplish a collaborative semantic task captured by a global co-safe Linear Temporal Logic (LTL) specification $\phi$. Similar to \cite{jones2013distribution,haesaert2018temporal,Kantaros2020perception}, this specification is defined over probabilistic atomic predicates that depend on both the multi-robot state $\bbp(t)$ and the estimate $\hat{\ccalM}(t)$. Specifically, we define perception-based predicates, defined as follows: 
\begin{equation}\label{eq:pip}
   \pi_p(\bbp(t),\hat{\ccalM}(t),\Delta)=
\begin{cases}
      \text{true}, & \text{if $p(\bbp(t),\hat{\ccalM}(t),\Delta)\geq 0$}\\
      \text{false}, & \text{otherwise}
    \end{cases} 
\end{equation}
In \eqref{eq:pip}, $\Delta$ is a set of user-specified and case-specific parameters (e.g., probabilistic thresholds or robots indices) and $p(\bbp(t),\hat{\ccalM}(t),\Delta): \Omega^N\times\hat{\ccalM}(t)\times\Delta\rightarrow\mathbb{R}$. Hereafter, when it is clear from the context, we simply denote a perception-based predicate by $\pi_p$.
First, we define the following function reasoning about the metric uncertainty:
\begin{align}
&p(\bbp(t),\hat{\ccalM}(t), \{j,\ell_i,r,\delta\})=\nonumber\\&\mathbb{P}(||\bbp_j(t)-\bbx_i(t)||\leq r)- (1-\delta).\label{ap1} 
\end{align}
The predicate associated with \eqref{ap1} is true at time $t$ if the probability of robot $j$ being within distance less than $r$ from target $\ell_i$ (regardless of its class) is greater than $1-\delta$, after applying control actions $\bbu_{0:t}$, for some user-specified parameters $r,\delta>0$. We also define the following function reasoning about both the metric and the semantic uncertainty:
%
\begin{align}
   &p(\bbp(t),\hat{\ccalM}(t),\{j,r,\delta,c\})=\nonumber\\&\max_{\ell_i}[\mathbb{P}(||\bbp_j(t)-\bbx_i(t)||\leq r)d_i(c)]-( 1-\delta).\label{apMS}
\end{align}
\normalsize
In words, the predicate associated with \eqref{apMS} is true at time $t$ if the probability of robot $j$ being within distance less than $r$ from at least one target with class $c$ is greater than $1-\delta$.

The syntax of co-safe LTL over a sequence of multi-robot states $\bbp(t)$ and uncertain semantic estimates $\hat{\ccalM}(t)$ is defined as $\phi::=\text{true}~|~\pi_p~|~\neg\pi_p~|~\phi_1\wedge\phi_2~|~\phi_1\vee\phi_2~|~\phi_1~\mathcal{U}~\phi_2,$
where 
\textcolor{black}{(i) $\pi_p$ is a perception-based predicate defined before} and (ii) $\wedge$, $\vee$, $\neg$, and $\mathcal{U}$, denote the conjunction, disjunction, negation, and until operator, respectively. Using $\ccalU$, the eventually operator $\Diamond$, can be defined as well \cite{baier2008principles}. 


\vspace{-0.2cm}
\subsection{Safe Planning over Uncertain Dynamic Semantic Maps}
\vspace{-0.1cm}
Given a task $\phi$, the sensing model \eqref{eq:measModelR}, the robot dynamics, and under assumptions (a)-(e), our goal is to select a stopping horizon $H$ and a sequence $\bbu_{0:H}$ of control inputs $\bbu(t)$, for all $t\in\set{0,\dots,H}$, that satisfy $\phi$ while minimizing a user-specified motion cost function. This gives rise to the following optimal control problem:

\vspace{-0.5cm}
\begin{subequations}
\label{eq:Prob2}
\begin{align}
& \min_{\substack{H, \bbu_{0:H}}} \left[J(H,\bbu_{0:H}) = \sum_{t=0}^{H}  c(\bbp(t),\bbp(t+1)) \right] \label{obj2}\\
& \ \ \ \ \ \ \  [\bbp_{0:H},\hat{\ccalM}_{0:H}]\models\phi  \label{constr1b} \\
& \ \ \ \ \ \ \   \bbp(t) \in \Omega_{\text{free}}^N, \label{obsFree}\\
& \ \ \ \ \ \ \   \bbp(t+1) = \bbf(\bbp(t),\bbu(t)) \label{constr3b}\\
& \ \ \ \ \ \ \ \hat{\bbx}(t+1)=A\hat{\bbx}(t)+B\boldsymbol\mu(t) \label{constr6b}\\
&  \ \ \ \ \ \ \  \Sigma(t+1)=\rho(\Sigma(t),\bbp(t+1)) \label{constr7b}\\
&  \ \ \ \ \ \ \  d(t+1)=d(0) \label{constr8}
\end{align}
\end{subequations}
\normalsize
where the constraints \eqref{constr3b}-\eqref{constr8} hold for all  $t\in[0,H]$. In \eqref{obj2}, any motion cost function $c(\bbp(t),\bbp(t+1))$ can be used associated with the transition cost from $\bbp(t)$ to $\bbp(t+1))$ as long as it is positive (e.g., traveled distance). The constraints \eqref{constr1b}-\eqref{obsFree} require the robots to accomplish the mission specification $\phi$ and always avoid the known obstacles/walls respectively. With slight abuse of notation, in \eqref{constr1b}, $[\bbp_{0:H},\hat{\ccalM}_{0:H}]$ denotes a finite sequence of length/horizon $H$ of multi-robot states and semantic estimates while $[\bbp_{0:H},\hat{\ccalM}_{0:H}]\models\phi$ means that the symbols generated along this finite sequence satisfy $\phi$. 
The constraint \eqref{constr3b} requires the robots to move according to their known dynamics. Also,  \eqref{constr6b}-\eqref{constr8} capture the offline map prediction (Section \ref{sec:updateMap}). 
\begin{problem} \label{prob}
Under assumptions (a)-(e), and given  an initial robot state $\bbp(0)$, a sensor model \eqref{eq:measModelR}, and a task $\phi$, compute a horizon $H$ and control inputs $\bbu(t)$ for all $t\in\{0,\dots,H\}$ as per \eqref{eq:Prob2}.
\end{problem}


\begin{rem}[Online Re-planning]\label{rem:objRec}
The object recognition method \eqref{eq:measModelC} is not required to solve \eqref{eq:Prob2} as, therein, the semantic uncertainty is not updated. In fact, \eqref{eq:Prob2} is an offline problem yielding open-loop/offline paths that are agnostic to \eqref{eq:measModelC}. A sampling-based algorithm to solve \eqref{eq:Prob2} is presented in Sections \ref{sec:reach}-\ref{sec:sampling}. In  Section \ref{sec:replan} we discuss how and when the offline paths may need to be revised online to adapt to perceptual feedback captured by \eqref{eq:measModelR}-\eqref{eq:measModelC}.
\end{rem}

%% file: files/alg_ICRA.tex
In this section, we present an algorithm to solve the semantic planning problem defined in \eqref{eq:Prob2}. To solve it, first, in Section \ref{sec:reach},  we convert \eqref{eq:Prob2} into a reachability problem that is defined over a hybrid state space. This state space consists of the multi-robot states, future states of the mobile semantic targets along with their corresponding quantified metric and semantic uncertainty (captured by Gaussian and discrete distributions), and a discrete automaton state space associated with the LTL task.
To solve this reachability problem, building upon our previous work \cite{Kantaros2020perception}, we propose a sampling-based algorithm; see Section \ref{sec:sampling}. We note that the major difference with \cite{Kantaros2020perception} lies in the structure of the state space that needs to be explored. Specifically, unlike \cite{Kantaros2020perception}, here, exploration of the future uncertain states of the dynamic targets is needed due to \eqref{constr6b}. In Section \ref{sec:replan}, we show how the proposed algorithm can be used online to to adapt to perceptual feedback \eqref{eq:measModelR}-\eqref{eq:measModelC}. In Section \ref{sec:extensions}, we discuss how to relax assumptions (a)-(e) (see Section \ref{sec:uncertainMap}).



\subsection{Reachability in Uncertain Hybrid Spaces}\label{sec:reach}
First, we convert \eqref{eq:Prob2} into a reachability problem. This is achieved by converting $\phi$ into a Deterministic Finite state Automaton (DFA), defined as follows \cite{baier2008principles}. 

\begin{definition}[DFA]
A Deterministic Finite state Automaton (DFA) $D$ over $\Sigma=2^{\mathcal{AP}}$ is defined as a tuple $D=\left(\ccalQ_{D}, q_{D}^0,\Sigma,\delta_D,q_F\right)$, where $\ccalQ_{D}$ is the set of states, $q_{D}^0\in\ccalQ_{D}$ is the initial state, $\Sigma$ is an alphabet, $\delta_D:\ccalQ_D\times\Sigma\rightarrow\ccalQ_D$ is a deterministic transition relation, and $q_F\in\ccalQ_{D}$ is the accepting/final state. 
\end{definition}

We also define a labeling function $L:\Omega^N\times\hat{\ccalM}(t)\rightarrow 2^{\mathcal{AP}}$ determining which atomic propositions are true given the current multi-robot state $\bbp(t)$ and the current map $\hat{\ccalM}(t)$. Given a robot trajectory $\bbp_{0:H}$ and a corresponding sequence of maps $\hat{\ccalM}_{0:H}$, we get the labeled sequence $L(\bbp_{0:H}, \hat{\ccalM}_{0:H}) =L([\bbp(0), \hat{\ccalM}(0)])\dots L([\bbp(H), \hat{\ccalM}(H)])$. This labeled sequence satisfies the specification $\phi$, if starting from the initial state $q_D^0$, each symbol/element in $L(\bbp_{0:H}, \hat{\ccalM}_{0:H})$ yields a DFA transition so that eventually -after $H$ DFA transitions- the final state $q_F$ is reached \cite{baier2008principles}. 
As a result, we can equivalently re-write \eqref{eq:Prob2} as follows:
\begin{subequations}
\label{eq:ProbR}
\begin{align}
& \min_{\substack{H, \bbu_{0:H}}} \left[J(H,\bbu_{0:H}) = \sum_{t=0}^{H}  c(\bbp(t),\bbp(t+1)) \right] \label{obj4}\\
& \ \ \ \ \ \ \  q_D(t+1) = \delta_D(q_D(t),\sigma(t)),\label{constr1c}\\
& \ \ \ \ \ \ \   \bbp(t+1) = \bbf(\bbp(t),\bbu(t)) \label{constr3c}\\
& \ \ \ \ \ \ \   \bbp(t) \in \Omega_{\text{free}}^N, \label{obsFree2}\\
& \ \ \ \ \ \ \ \hat{\bbx}(t+1)=\bbA\hat{\bbx}(t)+\bbB\boldsymbol\mu(t) \label{constr7x}\\
&  \ \ \ \ \ \ \  \Sigma(t+1)=\rho(\Sigma(t),\bbp(t+1)) \label{constr7c}\\
&  \ \ \ \ \ \ \  d(t+1)=d(0) \label{constr8c}\\
&  \ \ \ \ \ \ \  q_D(H)=q_F \label{constr9c}
\end{align}
\end{subequations}
where $q_D(0)=q_D^0$, $\sigma(t)=L([\bbp(t),\hat{\ccalM}(t)])$ and $\hat{\ccalM}(t)$ is determined by $\hat{\bbx}(t)$, $\Sigma(t)$, and $d(t)$. Note that the constraint in \eqref{constr1c} captures the automaton dynamics, i.e., the next DFA state that will be reached from the current DFA state under the observation/symbol $\sigma(t)$.
In words, \eqref{eq:ProbR}, is a reachability problem defined over a joint space consisting of the automaton state-space (see \eqref{constr1c}), the multi-robot motion space (see \eqref{constr3c}-\eqref{obsFree2}), and the future target states along with their corresponding metric and semantic uncertainty (see \eqref{constr7x}-\eqref{constr8c}) while the terminal constraint requires to reach the final automaton state (see \eqref{constr9c}).

\begin{algorithm}[t]\footnotesize
\caption{Safe Planning in Dynamic Semantic Maps}
\LinesNumbered
\label{alg:RRT}
\KwIn{ (i) maximum number of iterations $n_{\text{max}}$, (ii) robot dynamics \eqref{eq:rdynamics}, (iii) map distribution $\hat{\ccalM}(0)$, (iv) target dynamics $\bbg$, (v) initial robot configuration $\bbp(0)$, (vi) task $\phi$;}
\KwOut{Terminal horizon $H$, and control inputs $\bbu_{0:H}$}
Convert $\phi$ into a DFA\label{rrt:dfa}\;
Initialize $\ccalV = \set{\bbq(0)}$, $\ccalE = \emptyset$, $\ccalV_{1}=\set{\bbq(0)}$, $K_1=1$, and $\ccalX_g = \emptyset$\;\label{rrt:init}
\For{ $n = 1, \dots, n_{\text{max}}$}{\label{rrt:forn}
	 Sample a subset $\ccalV_{k_{\text{rand}}}$ from $f_{\ccalV}$\;\label{rrt:samplekrand}
     \For{$\bbq_{\text{rand}}(t)=[\bbp_{\text{rand}}(t),\hat{\ccalM}_{\text{rand}}(t),q_D]\in\ccalV_{k_{\text{rand}}}$}{\label{rrt:forq}
     Sample a control input $\bbu_{\text{new}}\in\mathit{U}$ from $f_{\mathit{U}}$\;\label{rrt:sampleu}
     $\bbp_{\text{new}}(t+1)=\bbf(\bbp_{\text{rand}}(t),\bbu_{\text{new}})$\;\label{rrt:pnew}
     \If{$\bbp_{\text{new}}(t+1)\in\Omega^N$}{\label{rrt:obsFree}
     $\hat{\bbx}_{\text{new}}(t+1)A = A(t)\hat{\bbx}_{\text{rand}}(t)+B(t)\mu_{\text{rand}}(t)$\;\label{rrt:xnew}
     $\Sigma_{\text{new}}(t+1)=\rho(\Sigma_{\text{rand}}(t+1),\bbp_{\text{new}}(t+1))$\;\label{rrt:Sigmaupdate}
     $d_{\text{new}}(t+1)=d(0)$\;\label{rrt:dnew}
     Construct map: $\hat{\ccalM}_{\text{new}}(t+1) = (\hat{\bbx}_{\text{new}}(t+1),\Sigma_{\text{new}}(t+1), d_{\text{new}}(t+1),g)$\;\label{rrt:updmap}
     Compute $q_D^{\text{new}}=\delta_D(q_D^{\text{rand}},L([\bbp_{\text{rand}}(t),\hat{\ccalM}_{\text{rand}}(t)]))$\label{rrt:feasTrans}\;
     \If{$\exists q_D^{\text{new}}$}{\label{rrt:feasTrans}
     Construct $\bbq_{\text{new}}=[\bbp_{\text{new}},\hat{\ccalM}_{\text{new}},q_D^{\text{new}}]$\;\label{rrt:qnew}
     Update set of nodes: $\ccalV= \ccalV\cup\{\bbq_{\text{new}}\}$\;\label{rrt:updV}
     Update set of edges: $\ccalE = \ccalE\cup\{(\bbq_{\text{rand}},\bbq_{\text{new}})\}$\;\label{rrt:updE}
     Compute cost of new state: $J_{\ccalG}(\bbq_{\text{new}})=J_{\ccalG}(\bbq_{\text{rand}})+c(\bbp_{\text{rand}},\bbp_{\text{new}})$\;\label{rrt:updCost}
     \If{$q_D^{\text{new}}=q_F$}{\label{rrt:updXg1}
     $\ccalX_g=\ccalX_g\cup\{\bbq_{\text{new}}\}$\;\label{rrt:updXg2}
     }
     Update the sets $\ccalV_{k}$\;\label{rrt:updVk}
     }}}
}
Among all nodes in $\ccalX_g$, find $\bbq_{\text{end}}(t_{\text{end}})$ \; \label{rrt:node}
$H=t_{\text{end}}$ and recover $\bbu_{0:H}$ by computing the path $\bbq_{0:t_{\text{end}}}= \bbq(0), \dots, \bbq(t_{\text{end}}) $\;\label{rrt:solution}
\end{algorithm}

\vspace{-0.2cm}
\subsection{\textcolor{black}{Sampling-based Algorithm}}\label{sec:sampling}
In this section, we present a sampling-based algorithm to solve \eqref{eq:ProbR}. The proposed algorithm  incrementally builds a tree that explores the hybrid space over which \eqref{eq:ProbR} is defined; see Alg. \ref{alg:RRT}.  
%
In what follows, we provide some intuition for the steps of Algorithm \ref{alg:RRT}. First, we denote the constructed tree by $\mathcal{G}=\{\mathcal{V},\mathcal{E},J_{\ccalG}\}$, where $\ccalV$ is the set of nodes and $\ccalE\subseteq \ccalV\times\ccalV$ denotes the set of edges. The set of nodes $\mathcal{V}$ contains states of the form $\bbq(t)=[\bbp(t), \hat{\ccalM}(t), q_D(t)]$, where $\bbp(t)\in\Omega$ and $q_D(t)\in\ccalQ_D$.\footnote{Throughout the paper, when it is clear from the context, we drop the dependence of $\bbq(t)$ on $t$.} The function $J_{\ccalG}:\ccalV\rightarrow\mathbb{R}_{+}$ assigns the cost of reaching node $\bbq\in\mathcal{V}$ from the root of the tree. The root of the tree, denoted by $\bbq(0)$, is constructed so that it matches the initial robot state $\bbp(0)$, the initial semantic map $\hat{\ccalM}(0)$, and the initial DFA state, i.e., $\bbq(0)=[\bbp(0), \hat{\ccalM}(0), q_D^0]$. 
By convention the cost of the root $\bbq(0)$ is $J_{\ccalG}(\bbq(0)) = 0$, 
while the cost of a node $\bbq(t+1)\in\ccalV$, given its parent node $\bbq(t)\in\ccalV$, is computed as 
\begin{equation}\label{eq:costUpd}
J_{\ccalG}(\bbq(t+1))= J_{\ccalG}(\bbq(t)) +  c(\bbp(t),\bbp(t+1)).
\end{equation}
Observe that by applying \eqref{eq:costUpd} recursively, we get that $J_{\ccalG}(\bbq(t+1)) = J(t+1,\bbu_{0:t+1})$ which is the objective function in \eqref{eq:Prob2}.

The tree $\ccalG$ is initialized so that $\ccalV=\{\bbq(0)\}$, $\ccalE=\emptyset$, and $J_{\ccalG}(\bbq(0)) = 0$ [line \ref{rrt:init}, Alg. \ref{alg:RRT}]. Also, the tree is built incrementally by adding new states $\bbq_\text{new}$ to $\ccalV$ and corresponding edges to $\ccalE$, at every iteration $n$ of Algorithm \ref{alg:RRT}, based on a \textit{sampling} [lines \ref{rrt:samplekrand}-\ref{rrt:sampleu}, Alg. \ref{alg:RRT}] and \textit{extending-the-tree} operation [lines \ref{rrt:pnew}-\ref{rrt:updVk}, Alg. \ref{alg:RRT}]. 
After taking $n_{\text{max}}\geq 0$ samples, where $n_{\text{max}}$ is user-specified, Algorithm \ref{alg:RRT} terminates and returns a feasible solution to \eqref{eq:ProbR} (if it has been found), i.e., a terminal horizon $H$ and a sequence of control inputs $\bbu_{0:H}$.

To extract such a solution, we need first to define the set $\ccalX_g\subseteq\ccalV$ that collects all states $\bbq(t)=[\bbp(t), \hat{\ccalM}(t), q_D(t)]\in\ccalV$ of the tree that satisfy 
$q_D(t)=q_F$, which captures the terminal constraint \eqref{constr9c} [lines \ref{rrt:updXg1}-\ref{rrt:updXg2}, Alg. \ref{alg:RRT}]. Then, among all nodes $\ccalX_g$, we select the node $\bbq(t)\in\ccalX_g$, with the smallest cost $J_{\ccalG}(\bbq(t))$, denoted by $\bbq(t_{\text{end}})$ [line \ref{rrt:node}, Alg. \ref{alg:RRT}]. 
Then, the terminal horizon is $H=t_{\text{end}}$, and the control inputs $\bbu_{0:H}$ are recovered by computing the path $\bbq_{0:t_{\text{end}}}$ in $\ccalG$ that connects $\bbq(t_{\text{end}})$ to the root $\bbq(0)$, i.e., $\bbq_{0:t_{\text{end}}}= \bbq(0), \dots, \bbq(t_{\text{end}})$ [line \ref{rrt:solution}, Alg. \ref{alg:RRT}]. 
Satisfaction of the constraints in \eqref{eq:ProbR} is guaranteed by construction of $\ccalG$. In what follows, we describe the core operations of Algorithm \ref{alg:RRT}, `\textit{sample}' and `\textit{extend}' that are used to construct the tree $\ccalG$.

\subsubsection{Sampling Strategy}\label{sec:sample} 
A new state $\bbq_\text{new}(t+1) =[\bbp_{\text{new}}, \hat{\ccalM}_{\text{new}}, q_D^{\text{new}}]$ is sampled at each iteration of Algorithm \ref{alg:RRT}. This state is generated as follows. To construct the state $\bbp_{\text{new}}$, we first divide the set of nodes $\ccalV$ into a \textit{finite} number of sets, denoted by $\ccalV_{k}\subseteq\ccalV$, based on the robot state $\bbp$ and the DFA state $q_D$ that comprise the states $\bbq\in\ccalV$. Specifically, $\ccalV_{k}$ collects all states $\bbq\in\ccalV$ that share the same DFA state and the same robot state (or in practice, robot states that are very close to each other). 
%
By construction of $\ccalV_{k}$, we get that $\ccalV = \cup_{k=1}^{K_n}\{\ccalV_{k}\}$, where $K_n$ is the number of subsets $\ccalV_{k}$ at iteration $n$.
Also, notice that $K_n$ is finite for all iterations $n$, due to the finite number of available control inputs $\bbu$ and the finite DFA state-space. 
At iteration $n=1$ of Algorithm \ref{alg:RRT}, it holds that $K_1=1$, $\ccalV_1=\ccalV$ [line \ref{rrt:init}, Alg. \ref{alg:RRT}]. 
Second, given the sets $\ccalV_k$, we first sample from a given discrete mass function $f_{\ccalV}(k|\ccalV):\set{1,\dots,K_n}\rightarrow[0,1]$ an index $k\in\set{1,\dots,K_n}$ that points to the set $\ccalV_{k}$ [line \ref{rrt:samplekrand}, Alg. \ref{alg:RRT}]. The mass function $f_{\ccalV}(k|\ccalV)$ defines the probability of selecting the set $\ccalV_{k}$ at iteration $n$ of Algorithm \ref{alg:RRT} given the set $\ccalV$. 


Next, given the set $\ccalV_{k_{\text{rand}}}$ sampled from $f_{\ccalV}$, we perform the following steps for all $\bbq\in \ccalV_{k_{\text{rand}}}$. Specifically, given a state $\bbq_{\text{rand}}$, we sample a control input $\bbu_{\text{new}}\in\mathit{U}$ from a discrete mass function $f_{\mathit{U}}(\bbu):\mathit{U}\rightarrow [0,1]$ [line \ref{rrt:sampleu}, Alg. \ref{alg:RRT}]. 
Given a control input $\bbu_{\text{new}}$ sampled from $f_{\mathit{U}}$, we construct the state $\bbp_{\text{new}}$ as $\bbp_{\text{new}}=\bbf(\bbp_{\text{rand}},\bbu_{\text{new}})$ [line \ref{rrt:pnew}, Alg. \ref{alg:RRT}]. If $\bbp_{\text{new}}$ belongs to the obstacle-free space, as required by \eqref{obsFree}, then the `extend' operation follows [line \ref{rrt:obsFree}, Alg. \ref{alg:RRT}]. More details about how these mass functions can be defined can be found in \cite{Kantaros2020perception}.

\subsubsection{Extending the tree}\label{sec:extend}

To build incrementally a tree that explores the hybrid space of \eqref{eq:ProbR}, we need to append to $\bbp_\text{new}$ the corresponding semantic map $\hat{\ccalM}_{\text{new}}$ determined by the parameters $(\hat{\bbx}_{\text{new}},\Sigma_{\text{new}},d_{\text{new}},\bbg)$ and DFA state $q_D^{\text{new}}$ [lines \ref{rrt:xnew}-\ref{rrt:updmap}, Alg. \ref{alg:RRT}]. Particularly, $\hat{\ccalM}_{\text{new}}$ is constructed so that $\hat{\bbx}_{\text{new}}, {\Sigma}_{\text{new}}$ is the a priori expected position and a posteriori covariance matrix, respectively, of the targets as required in \eqref{constr7x} and \eqref{constr7c}, i.e.,  $$\hat{\bbx}_{\text{new}}(t+1)=\bbA\hat{\bbx}_{\text{rand}}(t)+\bbB\boldsymbol\mu(t),$$ and  $$\Sigma_{\text{new}}(t+1)=\rho(\Sigma_{\text{rand}}(t),\bbp_{\text{new}}(t+1)).$$ Also, we have that $$d_{\text{new}}=d(0),$$
as required in \eqref{constr8}. We note again that these computations are possible due to assumptions (a)-(e) made in Section \ref{sec:uncertainMap}.

%


Next, to construct the state $\bbq_{\text{new}}$ we append to $\bbp_\text{new}$ and $\hat{\ccalM}_{\text{new}}$ the DFA state $q_D^{\text{new}}$,
as required by \eqref{constr1c}, i.e.,
$$q_D^{\text{new}}=\delta_D(q_D^{\text{rand}},L([\bbp_{\text{rand}},\hat{\ccalM}_{\text{rand}}])).$$
In words, $q_D^{\text{new}}$ is the automaton state that can be reached from the parent automaton state $q_D^{\text{rand}}$ given the observation $L([\bbp_{\text{rand}},\hat{\ccalM}_{\text{rand}}])$. 
If such a DFA state does not exist, then this means the observation $L([\bbp_{\text{rand}},\hat{\ccalM}_{\text{rand}}])$ results in violating the LTL formula and this new sample is rejected.\footnote{Note that when an LTL formula is violated, either a next state $q_D^{\text{new}}$ does not exist, or it exists but it is a deadlock automaton state; this is toolbox-specific. For simplicity, here we assume the former case.} Otherwise, the state $\bbq_{\text{new}}=(\bbp_\text{new},\hat{\ccalM}_{\text{new}},q_D^{\text{new}})$ is constructed [line \ref{rrt:qnew}, Alg. \ref{alg:RRT}] which is then added to the tree.

Given a state $\bbq_{\text{new}}$, we update the set of nodes and edges of the tree as $\ccalV = \ccalV\cup\{\bbq_{\text{new}}\}$ and $\ccalE = \ccalE\cup\{(\bbq_{\text{rand}},\bbq_{\text{new}})\}$, respectively [lines \ref{rrt:updV}-\ref{rrt:updE}, Alg. \ref{alg:RRT}]. The cost of the new node $\bbq_{\text{new}}$ is computed as 
$J_{\ccalG}(\bbq_{\text{new}})=J_{\ccalG}(\bbq_{\text{rand}})+c(\bbp_{\text{rand}},\bbp_{\text{new}})$ [line \ref{rrt:updCost}, Alg. \ref{alg:RRT}]. Finally, the sets $\ccalV_{k}$ are updated, so that if there already exists a subset $\ccalV_k$ associated with both the DFA state  $\bbq_D^{\text{new}}$ and the robot state $\bbp_{\text{new}}$, then $\ccalV_{k}=\ccalV_{k}\cup\set{\bbq_{\text{new}}}$. Otherwise, a new set $\ccalV_{k}$ is created, i.e., $K_n = K_n +1$ and $\ccalV_{K_n}=\set{\bbq_{\text{new}}}$ [line \ref{rrt:updVk}, Alg. \ref{alg:RRT}]. Recall that this process is repeated for all states  $\bbq_{\text{rand}}\in\ccalV_{k_{\text{rand}}}$ [line \ref{rrt:forq}, Alg.  \ref{alg:RRT}]. 


The proposed algorithm is probabilistically complete (Theorem \ref{thm:probCompl}) and asymptotically optimal (Theorem \ref{thm:asOpt}). The proofs of these results follow the same logic as in \cite{Kantaros2020perception} and, therefore, they are omitted. Particularly, the key idea is to show that the sampling-based algorithm exhaustively searches all possible sequences of control inputs $\bbu_{0:H}$ and finite horizons $H$.

\begin{theorem}[Probabilistic Completeness]\label{thm:probCompl}
If there exists a solution to \eqref{eq:Prob2}, then Algorithm \ref{alg:RRT} is probabilistically complete, i.e., the probability of finding a feasible solution, i.e., a feasible horizon $H$ and a feasible sequence of control inputs $\bbu_{0:H}$ for \eqref{eq:Prob2}, goes to $1$ as $n\to\infty$.
\end{theorem}

\begin{theorem}[Asymptotic Optimality]\label{thm:asOpt}
Assume that there exists an optimal solution to  \eqref{eq:Prob2}. Then, Algorithm \ref{alg:RRT} is asymptotically optimal, i.e., the optimal horizon $H$ and the optimal sequence of control inputs $\bbu_{0:H}$ will be found with probability $1$, as $n\to\infty$. In other words, the path generated by Algorithm \ref{alg:RRT} satisfies
$\mathbb{P}\left(\left\{\lim_{n\to\infty} J(H,\bbu_{0:H})=J^*\right\}\right)=1,$
where $J$ is the objective function of \eqref{eq:Prob2} and $J^*$ is the optimal cost.
\end{theorem}

\subsection{Online Execution and Re-planning}\label{sec:replan}

The proposed algorithm generates an open-loop sequence $\bbq_{0:H}=\bbq(0),\bbq(1),\dots,\bbq(H)$, where $\bbq(t)=[\bbp(t),\hat{\ccalM}(t),q_D(t)]$, so that the resulting robot trajectory $\bbp_{0:H}$ and sequence of maps $\hat{\ccalM}_{0:H}$ satisfy $\phi$. While executing these paths, the robots take measurements (see \eqref{eq:measModelR}-\eqref{eq:measModelC}) to update the (i) a posteriori mean and covariance of the landmarks and (ii) the discrete distribution associated with the landmark classes yielding an updated semantic map denoted by $\hat{\ccalM}_{\text{online}}(t)$ which may be different from the offline estimate of the map $\hat{\ccalM}(t)$. 
This may require the robots to replan to adapt to the new map. Formally, at time $t$ the robots replan if, for some $H-t>T>0$, the DFA state $q_D(t+T)$ in $\bbq_{0:H}$ cannot be reached, given the remaining robot path $\bbp_{t:t+T}$ and the sequence of maps $\hat{\ccalM}_{t:t+T}$, where $\hat{\ccalM}(t)=\hat{\ccalM}_{\text{online}}(t)$ and $\hat{\ccalM}(t+k)$, for all $k\in\{t+1,\dots,t+T\}$ is computed as in Section \ref{sec:updateMap}. Notice that as $T$ increases, the computational cost of reasoning whether replanning is needed increases as well. But
, a larger $T$ may prevent unnecessary replanning events. 
Re-planning is called every time when needed, till the robots complete the mission by reaching the final accepting state in the DFA.

\subsection{Extensions \& Future Work}\label{sec:extensions}
We discuss how assumptions (a)-(e) can be relaxed; see Section \ref{sec:PF}. Assumptions (a) and (e) require sensor and target linear models which may not hold in practice. This was required to compute offline the optimal a posteriori covariance matrices. This can be relaxed by linearizing them and applying an extended KF as e.g., in \cite{kantaros2021sampling}. Assumption (b) requires knowledge of the target dynamics including their control inputs. This is required to predict the future target states. This assumption can be relaxed by leveraging recurrent neural networks (RNNs) along with estimates of their predictive uncertainty \cite{alaa2020frequentist}. 
At test time, RNNs take as input the current trajectory of a target and predict its future waypoints along with corresponding confidence intervals \cite{alaa2020frequentist}. These confidence intervals along with the discrete distribution $d$ can be used to reason about task satisfaction. 
Assumptions (c) and (d) require prior information about the targets, so that initial paths can be designed. These can be relaxed by leveraging recently proposed exploration methods that can predict regions where semantic objects may be located in, based on the environmental context \cite{georgakis2021learning,georgakis2022uncertainty}. This will also allow us to handle cases where any available prior information is utterly wrong. 
Our future work will focus on formally relaxing these assumptions. 

%% file: files/sim_new.tex

In this section, we present experiments with aerial vehicles illustrating the performance of the proposed algorithm. In Section \ref{sec:replanSim}, we describe the semantic environment and the robot perceptual skills. In Section \ref{sec:case1}, we demonstrate the effect of the number of targets on the execution runtimes and the re-planning frequency. In Sections \ref{sec:case2}-\ref{sec:case3}, we illustrate the performance of the proposed algorithm as the metric and semantic priors become more inaccurate. Finally, in Section \ref{sec:case4&5}, we evaluate scalability of the algorithm with respect to the number $N$ of robots. 
Hereafter, to accelerate the construction of the trees, we employ the \textit{biased} sampling functions developed in \cite{Kantaros2020perception}.
%
Our experiments have been conducted on Gazebo (ROS, python3) on a computer with Intel Core i5 - 8350U 1.7GHz and 16Gb RAM.  Videos of the conducted experiments can be found in \cite{SimDynSemMaps}. 

\subsection{Experimental setup}\label{sec:replanSim}
\subsubsection{Robot Dynamics} 
Our experimental studies involve AsTech Firefly Unmanned Aerial Vehicles (UAVs) that operate over a semantic city with dimensions $150$m$\times 150$m. The UAVs are governed by first order linear dynamics where the UAV state includes the position, velocity, orientation, and biases in the measured angular velocities and acceleration; more details can be found in \cite{Furrer2016}. 
In general, the more complex the robot dynamics is, the longer it takes for sampling-based methods to generate feasible paths as they have to explore a larger state- and control- space. 
To mitigate this issue, an approach that we investigate in this section, is to generate paths for simple robot dynamics that need to be followed by robots with more complex dynamics. Specifically, in what follows, we use Algorithm \ref{alg:RRT} to synthesize  paths considering differential drive dynamics, defined in \eqref{eq:nonlinRbt}, that are simpler than the actual AsTech Firefly UAV dynamics. 

\footnotesize{
\begin{equation}\label{eq:nonlinRbt}
\begin{bmatrix}
   p_j^1(t+1) \\
  p_j^2(t+1)\\
  \theta_j(t+1)
  \end{bmatrix}=  
  \begin{bmatrix}
   p_j^1(t) \\
  p_j^2(t)\\
  \theta_j(t)
  \end{bmatrix}+    
 \begin{bmatrix}
   \nu (\text{sinc}(\tau\omega/2)\cos(\theta_j(t)+\tau\omega/2)) \\
   \nu (\text{sinc}(\tau\omega/2)\sin(\theta_j(t)+\tau\omega/2)) \\
  \tau \omega
  \end{bmatrix}
\end{equation}}
\normalsize
In \eqref{eq:nonlinRbt}, the robot state $\bbp_j(t)=[p_j^1(t),p_j^2(t),\theta_j]^T$ consists of both the position $[p_j^1(t),p_j^2(t)]$ and the orientation $\theta_j(t)$ of robot $j$ and $\tau$ is the sampling period. The available set of controls are $u\in\{0,1\}\text{m/s}$ and $\omega\in\set{0,\pm 1,\pm 2,\dots, \pm 180}\text{degree/s}$. Given the waypoints, generated by Algorithm \ref{alg:RRT}, we compute minimum snap trajectories that smoothly transition through all waypoints every $T=\tau$ seconds \cite{mellinger2011minimum}. This ensures that the drones can reach the waypoints at the same time instants that they would reach them if they were indeed governed by \eqref{eq:nonlinRbt}. 
The UAVs are controlled to follow the synthesized trajectories using the ROS package developed in \cite{Furrer2016}. 

%

\subsubsection{Semantic Targets}\label{sec:targetsSetup}
The environment is populated with $M$ semantic targets. Each target is assigned a class from the available classes $\ccalC = \{$'Parked car', 'Person', 'Enemy recon drone', 'Enemy security drone'$\}$ unless otherwise specified. 
Targets with classes 'Parked car' and 'Person' are stationary (i.e., $\bbx_i(t+1)=\bbx_i(t)$), while the rest are moving in the environment to accomplish their own individual tasks. We assume that these drones are governed by linear dynamics, as defined in \cite{Furrer2016}, and they apply control inputs $\boldsymbol\mu_i(t)$ to either move back and forth along a straight line or track a circular trajectory; see the videos in \cite{SimDynSemMaps}. The covariance matrix $\bbR_i$ of the noise $\bbv_i$ is selected so that it has zero non-diagonal entries and diagonal entries equal to 0.2.
%

\subsubsection{Perception System \eqref{eq:measModelR}-\eqref{eq:measModelC}}\label{sec:objRecDrone}
We assume that the drones are equipped with a downward facing sensor with square field of view with dimensions $24\times24$ m that can take noisy positional measurements of targets falling inside its field-of-view (see e.g., Figure \ref{fig:osc1} and  \cite{freundlich2018distributed}), i.e., the measurement of target $\ell_i$ by robot $j$ is:
\begin{equation}\label{eq:measPosDrone}
    \bby_{j,i}=\bbx_j + \bbv_j,
\end{equation}
where $\bbv_j$ is Gaussian noise with covariance matrix with diagonal entries equal to $2$ for all robots. As for the object recognition method \eqref{eq:measModelC}, we assume that our drones are equipped with a neural network that is capable of detecting objects and returning a discrete distribution over the available classes (see e.g. \cite{guo2017calibration}). 
In our simulations, we have simulated a neural network classifier with confusion matrix that has diagonal entries with values that range from $0.75$ to $0.9$; the remaining entries are generated randomly so that the sum of each row is equal to $1$. Any other model for the object detector in \eqref{eq:measModelC} can be used though.

\subsection{Effect of number of targets}\label{sec:case1}
\begin{figure}[t]
  \centering
%
  \subfigure[Planned path]{
    \label{fig:osc1}
  \includegraphics[width=0.47\linewidth]{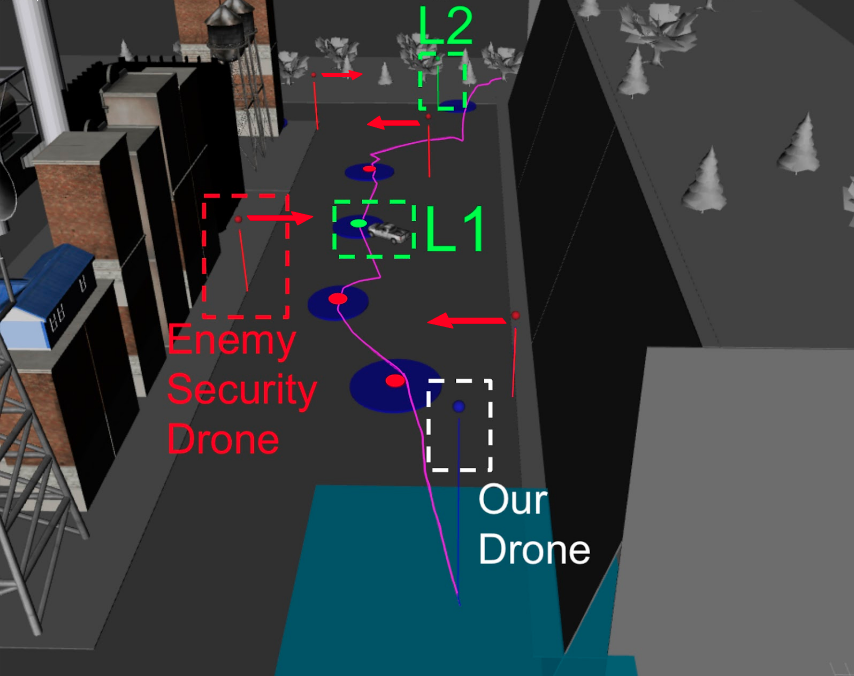}
  }
      \subfigure[Heading towards $\ell_1$ ('car') ]{
    \label{fig:osc2}
  \includegraphics[width=0.47\linewidth]{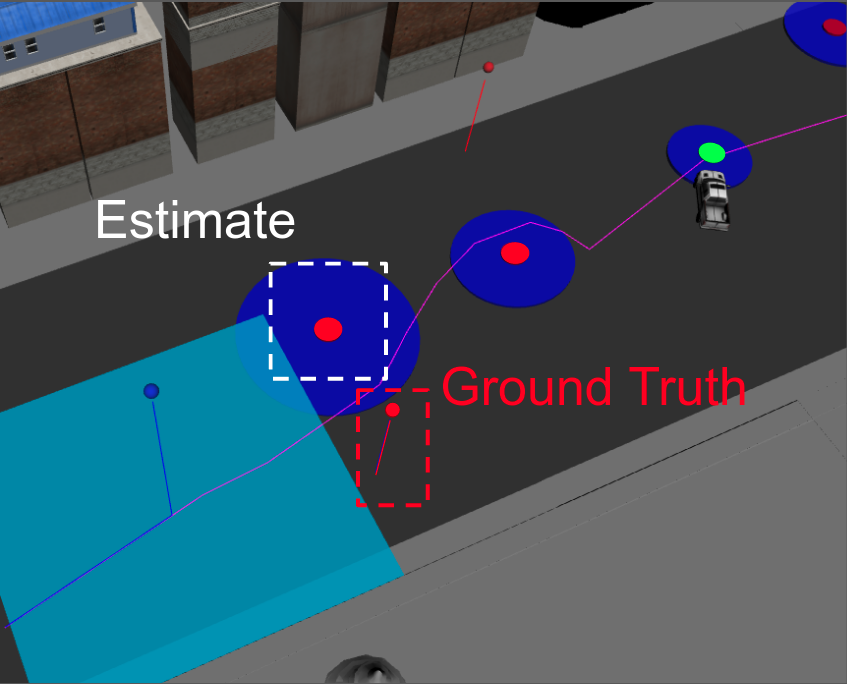}}
        \subfigure[Re-planning due to the new positional estimate of security drone]{
    \label{fig:osc3}
  \includegraphics[width=0.47\linewidth]{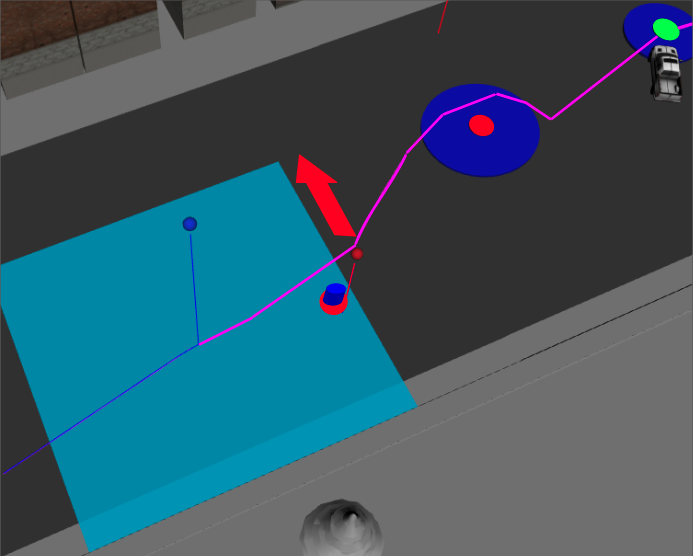}}
          \subfigure[New path generated]{
    \label{fig:osc4}
  \includegraphics[width=0.47\linewidth]{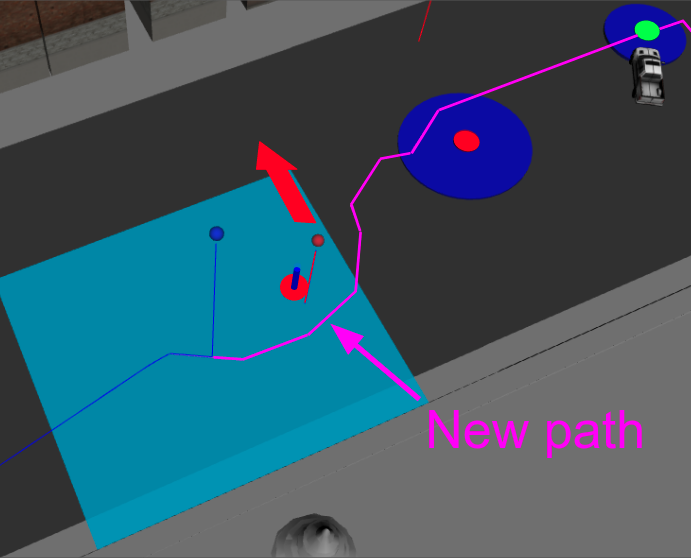}}
\subfigure[Following new path]{
    \label{fig:osc5}
  \includegraphics[width=0.47\linewidth]{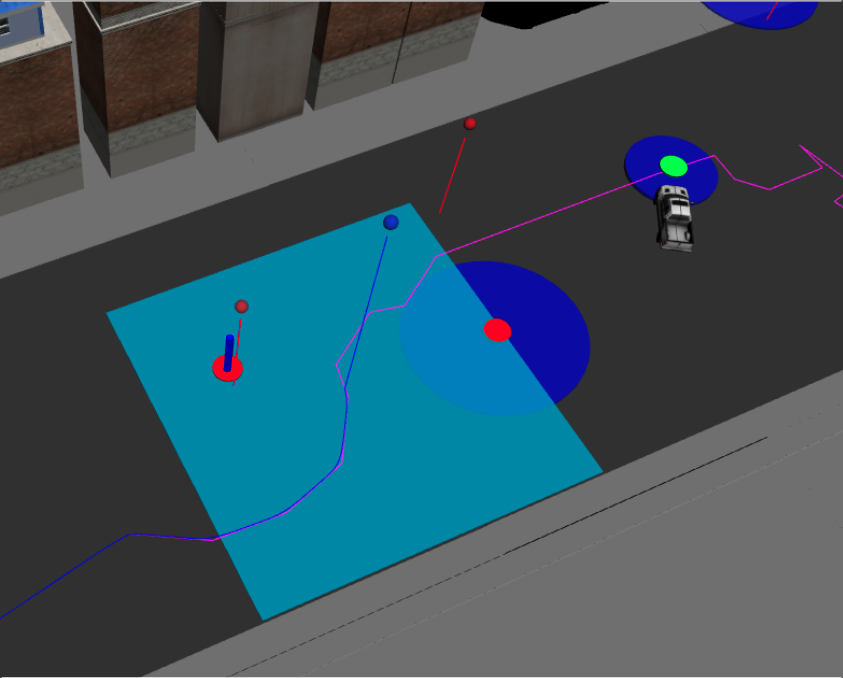}}
  \subfigure[Mission accomplished]{
    \label{fig:osc6}
  \includegraphics[width=0.47\linewidth]{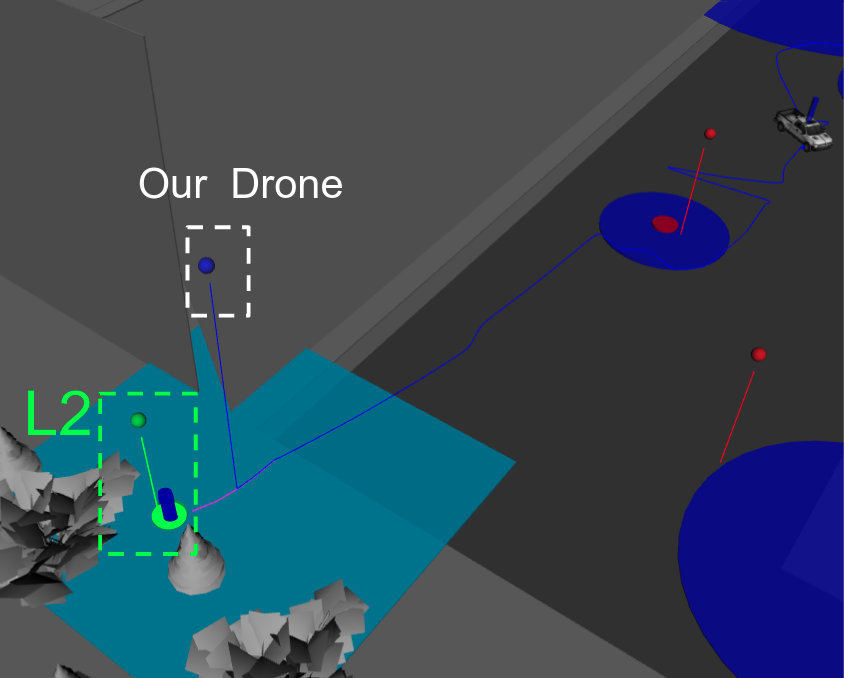}}  
 %
 \caption{Snapshots of a drone navigating an environment towards accomplishing the task in \eqref{eq:taskEx1}. The targets $\ell_1$ (abandoned car) and $\ell_2$ (recon drone) that should be visited are highlighted with green markers. The mobile enemy security drones that should always be avoided are highlighted with red markers. 
 The blue ellipsoid denotes the covariance of the prior Gaussian distributions with means illustrated with red and green disks.
 The drone initially heads towards the $\ell_1$ (`abandoned car') (Fig. \ref{fig:osc1}). The enemy drone is following an oscillatory path, i.e., it travels $20$ meters up and then $20$ meters down at $3$m/s velocity. When the drone reaches position in Fig. \ref{fig:osc3}) it detects and updates its belief about a target whose most likely class is `security drone' that should always be avoided. Given this new positional estimate of this target, if our drone continues on the original path it will come close to the security drone and, therefore, it will violate the probabilistic avoidance requirements in \eqref{eq:taskEx1}. Thus, the drone re-plans a new path based on the new estimate of the enemy drone and its predicted motion. The new path that is generated is shown in Fig. \ref{fig:osc4}. The drone then continues following this new path as shown in Fig. \ref{fig:osc5}). Every time the drone observes the enemy drone or target $\ell_1,\ell_2$ to be in a significantly different position than the expected one, it replans its path. Once the static target $\ell_1$ and dynamic target $\ell_2$ are approached as per the imposed probabilistic requirements, the mission terminates (Fig. \ref{fig:osc6}). 
}  \label{fig:Moving_Target}
\end{figure}

\begin{figure}[t]
  \centering
\includegraphics[width=1\linewidth]{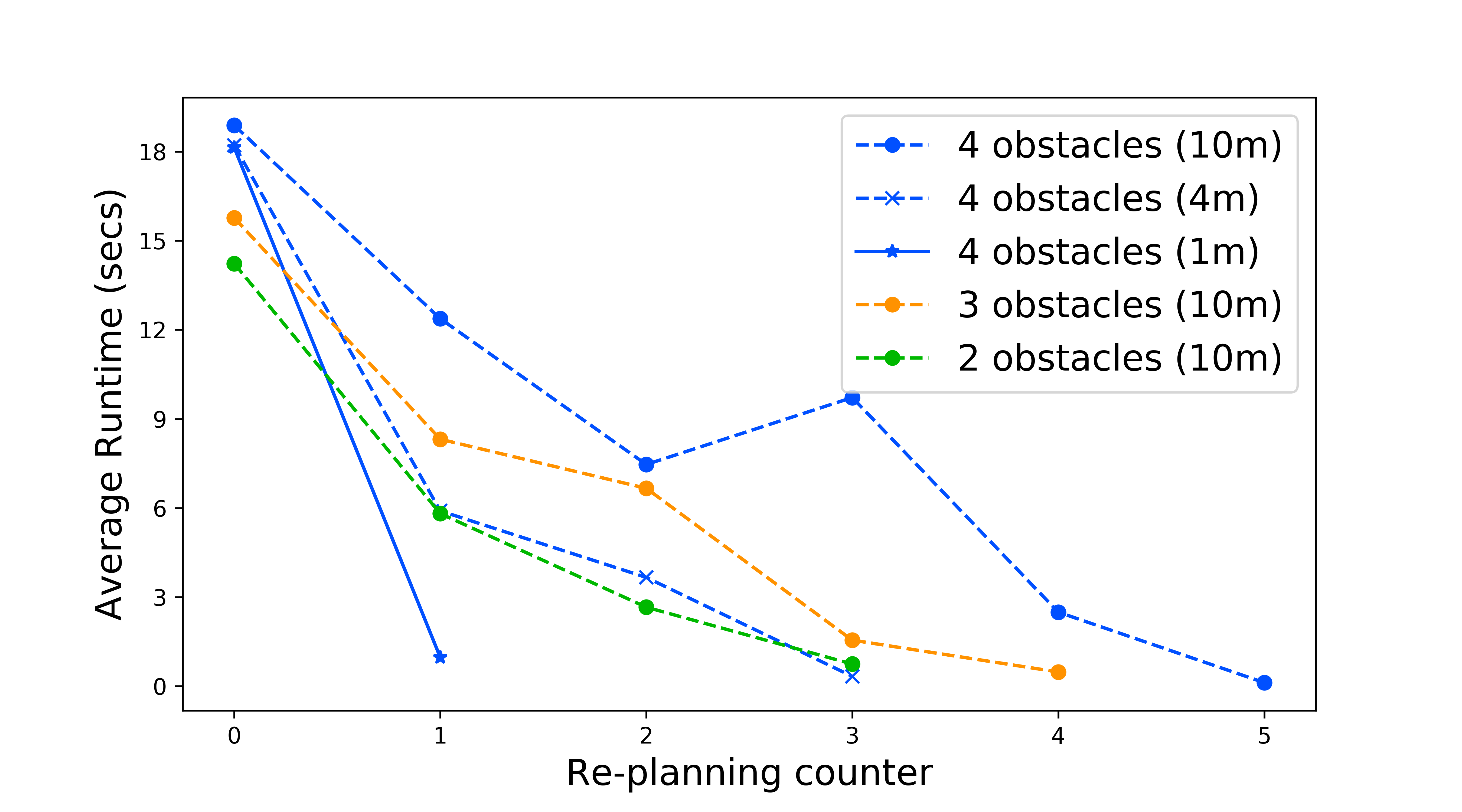}
    \caption{Graphical depiction of re-planning frequency and average re-planning run time for the case studies in Section \ref{sec:case1}-\ref{sec:case2}. Each curve corresponds to a different number of obstacles and different accuracy of prior estimates. For example, the topmost curve represents the setup with $4$ obstacles and poor prior estimate where the means are off by $10$ meters from ground truth. We observe that the number of times re-planning is called reduces as our prior estimate becomes accurate (see the three blue lines). Also, re-planning frequency increases with increasing obstacles in the environment (see green, yellow, blue lines). Note that run times when re-planning counter is $0$ refers to the time needed to design the initial path. Also note that these average runtimes are significantly smaller ($<1.5$ secs) when Algorithm \ref{alg:RRT} is not executed in the Gazebo simulation environment.} 
  \label{fig:graph-re-planning}
\end{figure}

In this section, we consider a case study where a single drone is responsible for accomplishing a spy/surveillance mission. Particularly, the drone must go into enemy territory and take photos of two specific targets, $\ell_1$ (an abandoned car) and $\ell_2$ (an enemy recon drone),  while avoiding any targets belonging to 'Enemy security drone' class, that might be in the way. All enemy drones fly at a altitude of $8$ meters, while the drone we control flies at at a fixed altitude of $16$ meters. For purpose of our simulation, we assume that the enemy security drones are equipped with cameras that allow them to detect any object that lies within a 3D ball of radius of $9$m. Thus, our drone has to always keep a distance of $9$m in the $xyz$ plane from them, or equivalently, $4$m in the $xy$ plane since all drones fly at fixed altitudes.
We can express this mission using the following co-safe LTL formula:  
\begin{align}\label{eq:taskEx1}
    \phi = &\Diamond[\pi_p(\bbp(t),\hat{\ccalM}(t),\{1, \ell_1,r_1,\delta_1\})\\&\wedge\Diamond\pi_p(\bbp(t),\hat{\ccalM}(t),\{1,\ell_2,r_2,\delta_2\})]\nonumber\\&\wedge[ \neg\pi_p(\bbp(t),\hat{\ccalM}(t),\{1,r_3,\delta_3,\text{`Security'}\})\nonumber\\&\ccalU \pi_p(\bbp(t),\hat{\ccalM}(t),\{1,\ell_2,r_2,\delta_2\})]\nonumber\\
    &\wedge[ \neg\pi_p(\bbp(t),\hat{\ccalM}(t),\{1,r_3,\delta_3,\text{`Security'}\})\nonumber\\&\ccalU \pi_p(\bbp(t),\hat{\ccalM}(t),\{1,\ell_1,r_1,\delta_1\})]\nonumber
\end{align}
where the atomic propositions $\pi_p$ are defined as in \eqref{ap1} and \eqref{apMS}, with parameters $r_1=r_2=2$ m, $r_3=4$ m, $\delta_1=\delta_2=0.25$, and $\delta_3=0.9$; all distances are computed based on the $x,y$ coordinates. This LTL formula corresponds to a DFA with $6$ states. The task demands our robot to visit $\ell_1$ and $\ell_2$ in that specific order, regardless of their class, while avoiding all targets with class `security drone'. 
We evaluate the performance of the proposed algorithm when there are $E\in\{2,3,4\}$ enemy security drones. In all case studies, we assume that difference between the estimated initial states (i.e., $\hat{\bbx_i}(0)$) of all the targets and their ground truth is (i.e., ${\bbx_i}(0)$) is approximately $10$m. Their covariances are also large indicative of poor confidence in their estimates. These prior Gaussian distributions as well as the actual initial locations of all targets are shown in Fig. \ref{fig:osc1}.
We assume the discrete distributions $d_i$ for all targets are correct, i.e., the most likely class of $\ell_i$ according to $d_i$ is the correct one.  
We ran each experiment thrice. In Fig. \ref{fig:graph-re-planning} we report the average time to design the initial path as well the average runtime to design the revised path when re-planning is triggered. We observe that the number of times re-planning is called, depends on the number of targets in the environment. The number of times re-planning is called is 5.33, 4.33 and 3 times on average for $E=4, E=3$ and $E=2$ respectively. 
Snapshots of the robot trajectory when $E=4$ along with one sample instance when re-planning was triggered are provided and discussed in Fig. \ref{fig:Moving_Target}.
%

\subsection{Effect of Metric Uncertainty}\label{sec:case2}
In this section, we consider the same setup as in Section \ref{sec:case1} with $M=6$, i.e., with four enemy security drones and one static ('abandoned car') and one dynamic ('recon drone') target. Our goal is to illustrate the performance of the proposed algorithm for various prior Gaussian distributions.
%
Specifically, in this experiment, we consider three configurations. 
In the first one (same setup as Fig. \ref{fig:Moving_Target}), our estimated positions of all targets are off by on an average $10$m (as in Section \ref{sec:case1}). In the second one, our estimates are off by on an average $4$m and in the third one they are off by an average of $1$m. At the same time the covariances for the targets in each setup gets progressively smaller indicative of more confidence in the estimate. For instance, the diagonal entries of covariance matrix of $\ell_1$ in the first, second, and the third setup are $[4,4]$, $[2,2.1]$, and $[1,0.8]$, respectively. Observe that even if the initial covariances are large enough, the proposed algorithm designs informative paths that aim to actively decrease the metric uncertainty so that the imposed probabilistic requirements are satisfied; we note that this is well-studied property in the related active sensing literature \cite{atanasov2015decentralized,kantaros2021sampling}.
We ran each setup thrice and counted the number of times re-planning was called. On an average the algorithm re-planned its path $5.33$ times in the first setup, $3$ times in the second setup and $1.33$ times in the last setup. This is in line with the intuition that better priors would result in `better' offline paths and, therefore, less frequent re-planning. The runtimes to design paths for the first run of each setup are shown in Fig. \ref{fig:graph-re-planning}; observe that paths can be designed quite fast $(<1.5\text{secs})$ 

\begin{figure}[t]
  \centering
\includegraphics[width=1\linewidth]{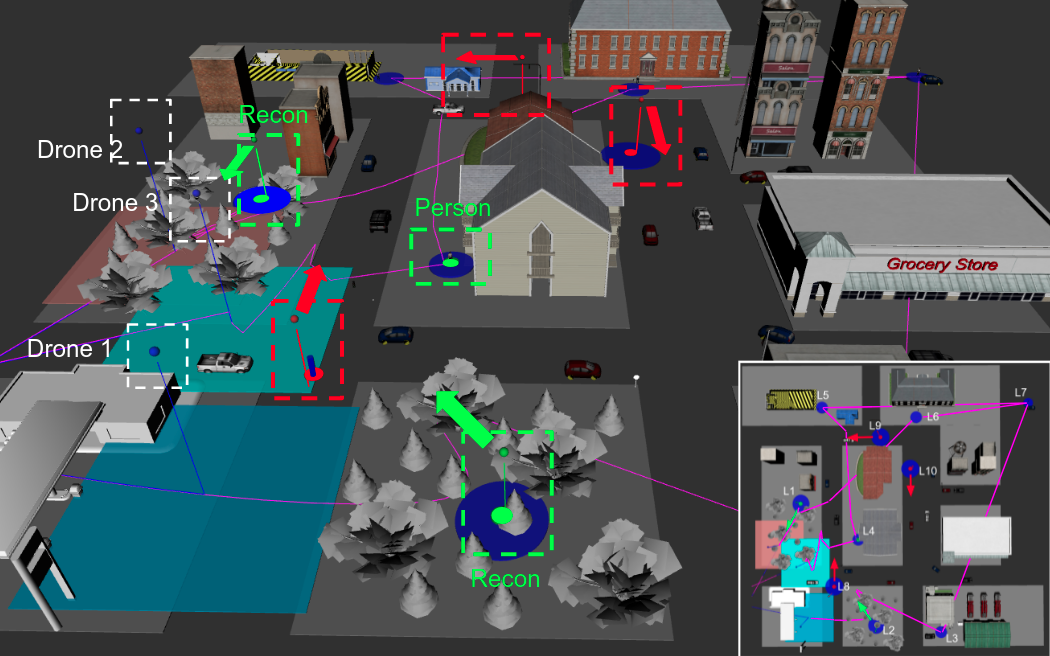}
\caption{\textcolor{black}{Graphical depiction the environment considered in 
Section \ref{sec:case4&5}. The colored squares represent the field of view of our drones. Target (recon drones) movement is shown with green arrows and security drone movement is shown with red arrows.}} 
  \label{fig:multibot2}
\end{figure}

\vspace{-0.5cm}
\textcolor{black}{\subsection{Effect of Semantic Uncertainty}\label{sec:case3}}
\begin{figure}[t]
  \centering
  \subfigure[Planned path]{
    \label{fig:class1}
  \includegraphics[width=0.9\linewidth]{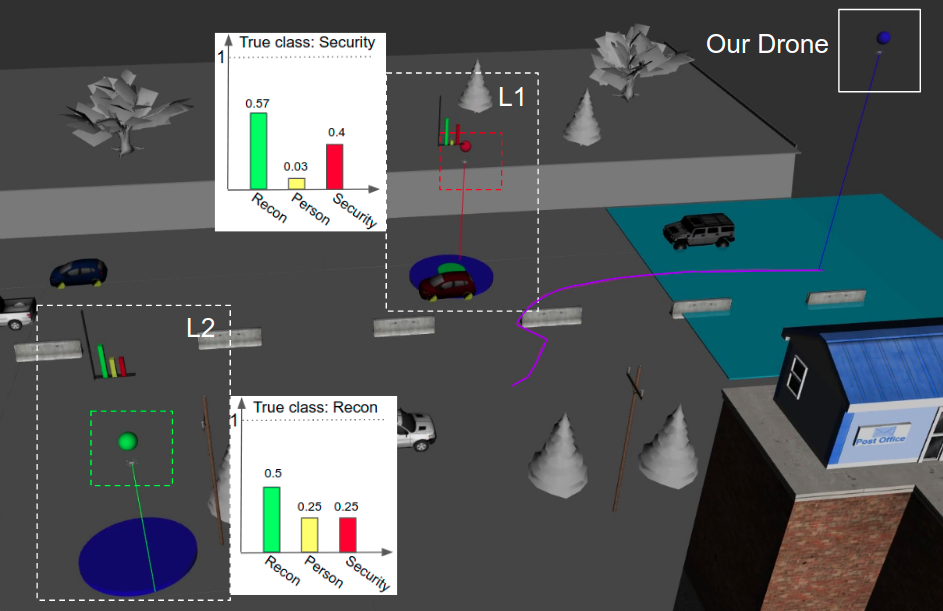}
  }
  \subfigure[Re-planned path ]{
    \label{fig:class2}
  \includegraphics[width=0.9\linewidth]{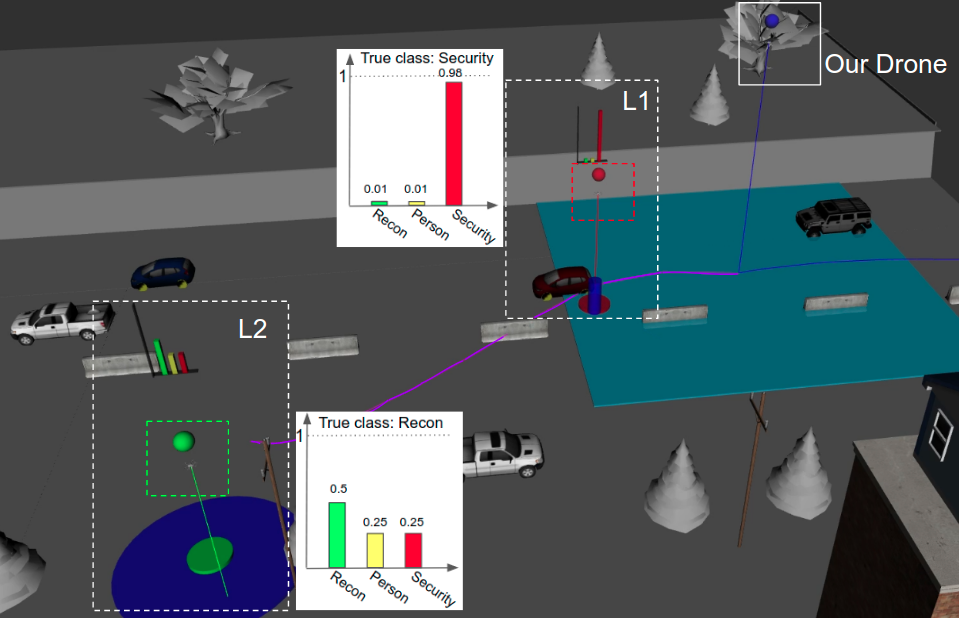}}
 \caption{Snapshots of a drone navigating an environment to accomplish \eqref{eq:taskEx2}. Each target is illustrated by a red or green sphere depending on their actual label: red for `security' and green for `recon'. Fig. \ref{fig:class1} shows the initial path. Due to the incorrect semantic prior, our robot plans a path towards $\ell_1$ as it believed to be a `recon drone' (see $d_1$) and, therefore, it has to be approached.
 However, as the drone follows the designed path, it updates its belief about the semantic label of $\ell_1$, which is now correctly classified as `security drone' that should always be avoided. Thus, the drone designs a new path towards $\ell_2$ that is (correctly) expected to be a recon drone; see Fig. \ref{fig:class2}.}
  \label{fig:Incorrect_classification}
\end{figure}
In this section, our goal is to illustrate the performance of the proposed algorithm for various prior discrete distributions modeling semantic uncertainty. To demonstrate this, we consider an environment with $M=2$ targets and a set of classes defined as  $\ccalC = \{$'Enemy recon drone', 'Person', 'Enemy security drone'$\}$. Our goal is to control a single drone so that it achieves a task requiring to eventually find and approach a target with class `recon drone', while always avoiding all targets with class `security drone'. This task is specified by the following LTL formula:
\begin{align}\label{eq:taskEx2}
    \phi = &[\Diamond\pi_p(\bbp(t),\hat{\ccalM}(t),\{1,r_1,\delta_1,\text{`Recon'}\})]\\&\wedge[ \neg\pi_p(\bbp(t),\hat{\ccalM}(t),\{1,r_2,\delta_2,\text{`Security'}\})\nonumber\\&\ccalU\pi_p(\bbp(t),\hat{\ccalM}(t),\{1,r_1,\delta_1,\text{`Recon'}\})]\nonumber
\end{align}
where the atomic propositions $\pi_p$ are defined as in 
\eqref{apMS}, with parameters $r_1=2$m, $r_2=4$m, $\delta_1=0.25$, and $\delta_2=0.9$. This LTL formula corresponds to a DFA with 3 states. The prior map is defined so that the discrete distribution for target $\ell_1$ is `incorrect'. Specifically, target $\ell_1$ is a `security drone' but its prior distribution is defined so that it is a `recon drone' with probability $0.57$ and a `security drone' with probability $0.4$. Also, target $\ell_2$ is a `recon drone' and its discrete distribution is defined so that it is a `recon drone' with probability $0.5$ and a `security drone' with probability $0.25$.
Observe that these semantic priors, besides being `incorrect', they are also not informative enough to satisfy the probabilistic requirements captured in \eqref{eq:taskEx2}. As a result, \eqref{eq:ProbR} is infeasible and, therefore, Alg. \ref{alg:RRT} cannot generate a path. To address this issue, in this case study, we aim to design least violating paths. To achieve this, when we build the tree to design the initial paths, instead of checking satisfaction of \eqref{apMS}, we check satisfaction of a relaxed version of it. Specifically, we deterministically assign to each target the most likely class determined by $d_i$. Thus, in this example, both $\ell_1$ and $\ell_2$ are (incorrectly) considered `recon drones'. Then, among all targets with a label $c$ (e.g., `recon drone'), we investigate satisfaction of the following predicate:
%
\begin{align}
    &p(\bbp(t),\hat{\ccalM}(t),\{j,r,\delta,c\})=\nonumber\\&\max_{\ell_i,~ c_i=c}[\mathbb{P}(||\bbp_j(t)-\bbx_i(t)||\leq r)]-( 1-\delta).\label{apMS_relax}
\end{align}
Given this semantic prior, the first feasible path generated by Alg. \ref{alg:RRT} leads the drone towards $\ell_1$; see Fig. \ref{fig:Incorrect_classification}. However, once the target is in sight, our robot updates $d_1$ and realizes that $\ell_1$ is in fact a security drone. At this point, given this new semantic belief, the robot designs a new path that leads to $\ell_2$, while avoiding $\ell_1$. Snapshots of this experiment are shown in Fig. \ref{fig:Incorrect_classification} and the simulation is included in \cite{SimDynSemMaps}. 
Note that, it is challenging to determine if the semantic priors are informative enough so that \eqref{eq:ProbR} is feasible. A possible approach to address this is to run the proposed algorithm with the original predicates for a large number of iterations. In parallel, the algorithm is ran with the relaxed predicates in case the original problem is infeasible. 
\textcolor{black}{\subsection{Multi-Robot Experiments}\label{sec:case4&5}}
In this section, to evaluate the performance of the proposed algorithm with respect to the number $N$ of robots, we consider multi-robot experiments in an environment with $M=12$ targets. An illustration of the considered environment is shown in Fig. \eqref{fig:multibot2}.  
We consider the LTL task:
\begin{equation}\label{eq:taskEx3}
    \phi=\Diamond(\xi_{1}\wedge\Diamond\xi_{2})\wedge\Diamond\xi_3\wedge\Diamond\xi_4\wedge\neg\xi_3 \ccalU \xi_1\wedge\neg\xi_4\ccalU \xi_2,
\end{equation}
where each $\xi_i$ is a Boolean formula defined over atomic predicates of the form \eqref{ap1} for various robots and landmarks. 
For example $\xi_1$ when $N=2$ is defined as 
\begin{align}\label{eq:taskEx4}
\xi_1 =  &\pi_p(\bbp(t),\hat{\ccalM}(t),\{1, \ell_1,r_1,\delta_1\})\\&\wedge\pi_p(\bbp(t),\hat{\ccalM}(t),\{2,\ell_2,r_2,\delta_2\})\nonumber
\end{align}
where the atomic propositions $\pi_p$ are defined as in \eqref{ap1} with parameters $r_1=r_2=2$m, $\delta_1=\delta_2=0.2$..
The LTL formula \eqref{eq:taskEx3} corresponds to a DFA with $14$ states.
For $N\in\{1,2,3,5,10\}$ robots, the time needed to design the initial feasible path is $\{0.01, 0.03,0.21,1.58,7.35\}$ mins, respectively; these runtimes depend on the DFA size and the environmental structure. Notice that as $N$ increases, these runtimes tend to increase. This may prevent application of the proposed algorithm for controlling large-scale multi-robot systems in rapidly-changing environments. A potential approach to mitigate this issue is to perform task decomposition, as e.g., in \cite{kantaros2020reactive} which, however, may sacrifice completeness.

